\pdfoutput=1
\documentclass[11pt]{article}

\usepackage[final]{acl}

\usepackage{times}
\usepackage{latexsym}
\usepackage[T1]{fontenc}
\usepackage[utf8]{inputenc}

\usepackage{microtype}
\usepackage{multirow}
\usepackage{booktabs}
\usepackage{inconsolata}

\usepackage{graphicx}

\usepackage{longtable}

\usepackage{color}
\definecolor{UserColor}{rgb}{0.6, 0.6, 0.6}
\usepackage{tcolorbox}
\tcbuselibrary{breakable}
\newtcolorbox{myblock}[1]{colback=UserColor!10, colframe=UserColor, coltitle=white, title=#1, fonttitle=\bfseries, breakable}

\usepackage[nottoc]{tocbibind}
\usepackage{minitoc}
\usepackage{bbm}

\newcommand{\ie}{\textit{i}.\textit{e}.,\ }
\newcommand{\eg}{\textit{e}.\textit{g}.,\ }

\newcommand{\myfig}[1]{Figure \ref{#1}}
\newcommand{\mytable}[1]{Table \ref{#1}}

\newcommand{\mysec}[1]{Section \ref{#1}}

\newcommand{\dotieconcat}[2]{% auxiliary macro, don't use it directly
  \text{\raisebox{.8ex}{$\smallfrown$}}%
}

\makeatletter
\newcommand\dslfontsize{\@setfontsize\dslfontsize\@viipt\@viiipt}
\renewcommand\scriptsize{\@setfontsize\subfigcap{7}{8}}%
\makeatother

\newcommand\blfootnote[1]{%
  \begingroup
  \begin{NoHyper}%
  \renewcommand\thefootnote{}\footnote{#1}%
  \addtocounter{footnote}{-1}%
  \end{NoHyper}%
  \endgroup
}

\usepackage{color}
\usepackage{xcolor}
\definecolor{codegray}{rgb}{-0.5,0.5,0.5}
\usepackage{makecell}

\usepackage[normalem]{ulem}

\title{Rethinking Creativity Evaluation: A Critical Analysis of Existing Creativity Evaluations}

\author{
 \textbf{Li-Chun Lu\textsuperscript{1$\dagger$}}
 \quad
 \textbf{Miri Liu\textsuperscript{2}}
 \quad
 \textbf{Pin-Chun Lu\textsuperscript{3}}
 \quad
 \textbf{Yufei Tian\textsuperscript{2}}
\\
 \textbf{Shao-Hua Sun\textsuperscript{1,4}}
 \quad
 \textbf{Nanyun Peng\textsuperscript{2}}
\\
\\
 \textsuperscript{1}Graduate Institute of Communication Engineering, National Taiwan University
 \\
 \textsuperscript{2}Computer Science Department, University of California, Los Angeles
 \\
 \textsuperscript{3}Graduate Institute of Networking and Multimedia, National Taiwan University
 \\
 \textsuperscript{4}Department of Electrical Engineering, National Taiwan University
\\
}

\begin{document}
\doparttoc 
\faketableofcontents

\maketitle
\begin{abstract}
We examine, analyze, and compare four representative creativity measures—perplexity, LLM-as-a-Judge, the Creativity Index (CI; measuring n-gram overlap with web corpora), and syntactic templates (detecting repetition of common part-of-speech patterns)—across the diverse creative domains, such as creative writing, unconventional problem-solving, and research ideation. For each domain, we compile datasets with human-aligned creative and uncreative examples and evaluate each metric’s ability to discriminate between the two sets. Our analyses reveal limited consistency both across domains and metrics, as metrics that distinguish creativity in one domain fail in others (\eg CI correctly distinguishes in creative writing but fails in problem-solving), and different metrics often disagree on the same data points (\eg CI suggests one set to be more creative, while perplexity indicates the other set to be more creative.) We highlight key limitations, such as perplexity reflecting fluency rather than novelty; LLM-as-a-Judge producing inconsistent judgments under minor prompt variations and exhibiting bias towards particular labels; CI primarily measuring lexical diversity, with high sensitivity to implementation choices; and syntactic templates being ineffective in settings dominated by formulaic language. Our findings underscore the need for more robust, generalizable evaluation frameworks that better align with human judgments of creativity. We release the datasets and evaluation code: \url{https://github.com/lichun-19/creative_eval}.

\blfootnote{\textsuperscript{$\dagger$} Work done during Li-Chun’s visit to UCLA.\\Correspondence to: Nanyun Peng \textless{}violetpeng@cs.ucla.edu\textgreater{}}

\end{abstract}

\section{Introduction}
\label{intro} 
Creativity is essential to human progress across art, daily problem-solving, and scientific discovery~\citep{cropley2006role, feist1998meta, simonton2004creativity}. Large language models (LLMs) perform well on structured tasks such as question answering and logical reasoning~\citep{goel2023, cabral2024, pu2023summarizationalmostdead, li2024flexkbqaflexiblellmpoweredframework}, but their creative abilities remain limited~\citep{Sawicki2023OnTP, sawicki2023bitsgrassdoesgpt, chakrabarty2024, mirowski}. As interest in machine creativity grows, so too does the need for reliable evaluation metrics that can effectively benchmark LLM performance across creative domains.

\begin{figure*}[t]
  \includegraphics[width=1\linewidth]{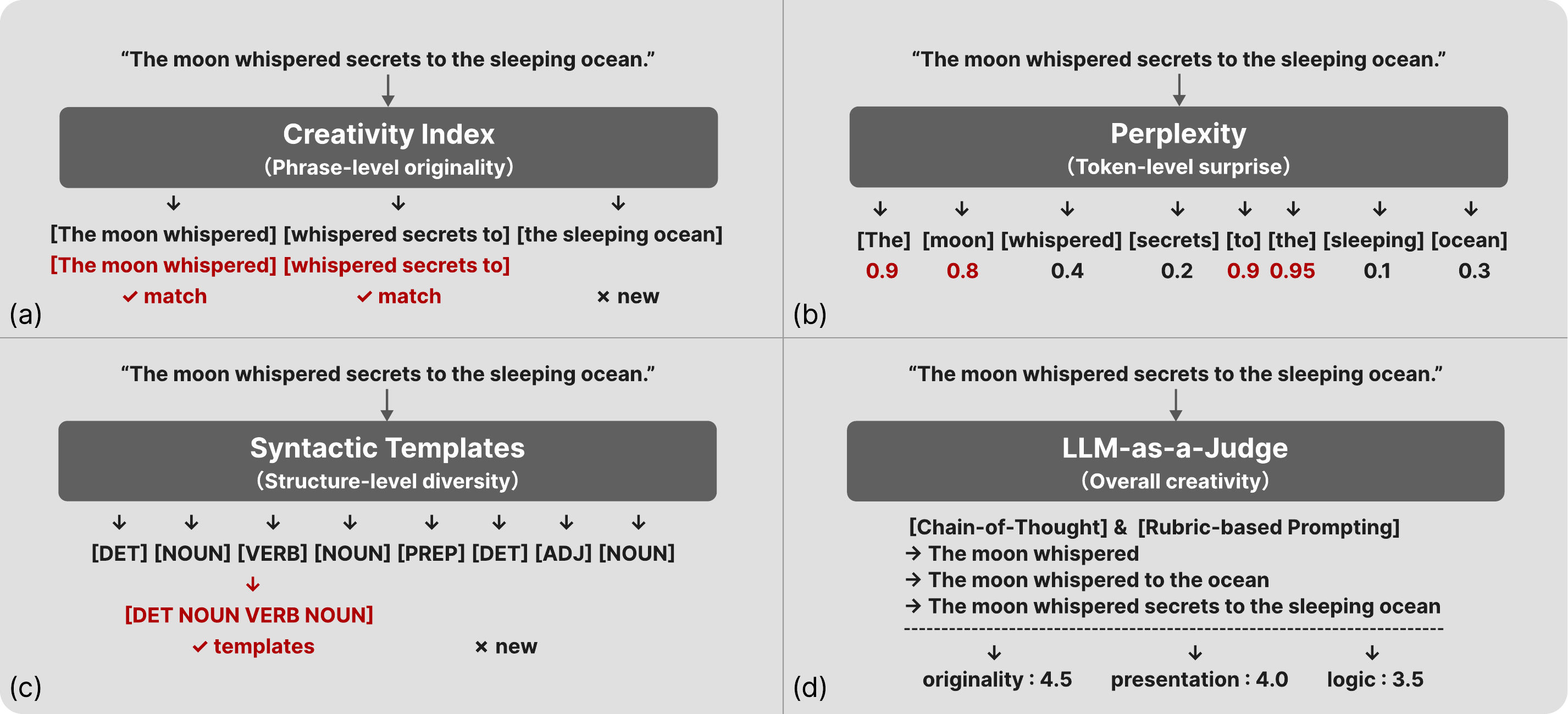}
  \caption{\textbf{Illustrations of creativity metrics selected for evaluation.}
(a) \textbf{Creativity Index} captures phrase-level originality by identifying reused segments from large-scale web datasets.
(b) \textbf{Perplexity} measures token-level unexpectedness via language model prediction probabilities.
(c) \textbf{Syntactic templates} evaluate structure-level novelty by detecting reliance on common syntactic patterns.
(d) \textbf{LLM-as-a-Judge} incorporates rubric-based scoring and chain-of-thought reasoning to assess overall creative quality.
}
\label{fig:metrics_intro}
\end{figure*}

The inherently multidimensional and domain-specific nature of creativity has led to a diverse range of evaluation methods. Prior work has proposed various automatic metrics to capture different aspects of creativity, such as semantic distance to measure novelty in text generation~\citep{zhao2019, haque2022}, and metrics like the Creativity Index and syntactic templates for assessing linguistic creativity~\citep{li2016, lu2025salieri, liu2024infinigramscalingunboundedngram, shaib2024}. LLMs have also been used as evaluators in tasks like the Alternative Uses Task (AUT) and creative writing~\citep{Goes2023, organisciak, distefano2024, luchini2025automated, luchini2025multilingual}. 
These methods highlight the diversity of creativity and the complexity of its evaluation. This raises an important question: how can we select evaluation metrics accurately reflecting human-perceived creativity across domains?

To address this question, we analyze four representative creativity metrics, each reflecting a different level of granularity in creativity evaluation as illustrated in~\myfig{fig:metrics_intro}. We adopt the commonly used definition of creativity~\citep{guerberof2022creativity, mukherjee2023creative, ismayilzada2025creative}, which treats creativity as the combination of \textit{novelty} and \textit{acceptability} (also referred to as quality or usefulness, \ie the idea is appropriate or fit for its purpose and constraints). In our setting, we emphasize novelty, as our dataset construction process already filters out infeasible or low-quality examples. We use the Creativity Index~\citep{lu2025salieri} to assess phrase-level creativity by computing n-gram overlap with large-scale web corpora, perplexity to capture token-level diversity, syntactic templates \citep{shaib2024} to evaluate structural creativity, and LLM-as-a-Judge for a holistic assessment~\citep{merrill2024evaluating, crossley2016idea, nguyen2024}. To evaluate the alignment between these metrics and human perceptions of creativity, we apply them across three domains where machine creativity has been extensively studied~\citep{ismayilzada2024creativityaiprogresseschallenges}: artistic creativity (\eg creative writing), unconventional problem-solving, and research ideation. 
For each domain, we curate a dataset of both creative and uncreative examples, reflecting human judgments, and assess each metric's performance within and across these domains.

Our analysis reveals that the four creativity metrics exhibit limited consistency both across domains and with one another. We further analyze each metric's behavior:
\begin{itemize}
    \item \textbf{Creativity Index} is weakened by its sensitivity to the choice of L-uniqueness and the selection of reference corpora, \eg difference between two cases can vary by a factor of two when using L = 5–7 versus L = 5–11.
    \item \textbf{Perplexity}, while associated with diversity and surprise~\citep{basu2021mirostat}, exhibits substantial overlap in creative and uncreative score distributions across all domains.
    \item \textbf{Syntactic templates} are weak in settings where novel ideas are often expressed through formulaic language, such as research ideation and problem-solving.
    \item \textbf{LLM-as-a-Judge} is highly prompt-sensitive and prone to one-sided bias across models; furthermore, repeated evaluations of the same instance by the same judge exhibit only 40\% consistency across three runs.
\end{itemize}
In conclusion, our findings reveal critical inconsistencies across metrics and domains, emphasizing the need for more robust, generalizable evaluation frameworks aligned with human creativity judgments. By clarifying the strengths and limitations of existing metrics, our study provides a foundation for future methods and informs current best practices in creativity evaluation.

\section{Background on creativity evaluation} 

As LLMs advance in generating creative content, the need for effective evaluation methods has grown. We review existing metrics and discuss their applicability to evaluating machine-generated content.

Classical psychological tests, including the Torrance Tests of Creative Thinking~\citep{torrance1966}, Divergent Association Task~\citep{olson2021}, Remote Associates Test~\citep{mednick1962associative}, and Alternative Uses Task~\citep{guilford1967}, capture different facets of creative thinking, but are limited for evaluating LLMs. Firstly, traditional metrics like fluency become less informative as LLMs generate extensive content with ease~\citep{guzik2023, lu2024llm}; secondly, these tests require significant human effort~\citep{torrance1966, olson2021, guilford1967}; and finally, subjective evaluation introduces potential biases~\citep{beaty2021}.

As LLMs are increasingly applied to creative domains such as story writing~\citep{fan2018, akoury2020, gomez2023} and music composition~\citep{ding2024songcomposerlargelanguagemodel, parker2024, yuan2024chatmusician}, several benchmarks have been introduced to assess their creative capacity~\citep{lu2024benchmarking, zhang2025noveltybench, atmakuru2024cs4}. These benchmarks provide system-level comparisons across models. Our focus, however, is on the level of individual outputs: developing automatic metrics that can assess creativity in open-ended settings beyond the constraints of predefined tasks.

Automated measures including linguistic diversity~\citep{li2016, lu2025salieri, liu2024infinigramscalingunboundedngram, shaib2024}, text perplexity~\citep{fan2018}, and LLM-based judgment~\citep{ruan2025, ye2024, lin2024, pai2026billy} have been proposed as scalable alternatives to human assessment. However, LLM-based evaluation is susceptible to biases such as positional bias and preference for model-generated outputs~\citep{shi2024, panickssery2024, ye2024bias}, and few works have examined whether these metrics meaningfully capture conceptual novelty, which is a gap we address in this study.

To close this gap, our framework analyzes creativity through a technical segmentation at the token, phrase, and overall levels, complementing prior work that draws on human judgments using adapted psychological tests~\citep{chakrabarty2024, tian2024a, lu2024llm} or qualitative frameworks emphasizing surprise, diversity, and novelty~\citep{ismayilzada2024evaluating, ismayilzada2025creative}. By systematically examining existing measures across diverse creative domains, we provide an interpretable foundation for systematic, automated analysis and guide the development of more effective assessment methods for future research.
\section{Core domains of creativity selected for evaluation}
To systematically analyze existing creativity measures across diverse creative domains, we introduce the domains we selected in~\mysec{sec:domains}, and describe the curation process in ~\mysec{sec:datasets}.

\subsection{Domains}
\label{sec:domains}
Creativity is an incredibly broad concept spanning multiple disciplines; to aid the evaluation of existing creativity metrics, we first identify the core domains of creativity. 

We follow the broad categorization by~\citet{ismayilzada2024creativityaiprogresseschallenges} and select three domains representing distinct aspects of creativity: \textbf{creative writing} (artistic and aesthetic approaches), \textbf{unconventional problem-solving} (functional thinking), and \textbf{research ideation} (logical synthesis of existing facts).

\textbf{Creative writing.} \textsc{Creative writing} serves as a representative sub-
field for artistic creativity, which involves generating original, aesthetically engaging content that elicits emotional or intellectual responses~\citep{jean2024creative, kaimal2017functional, fancourt2019how, kim2015community, felton2015widening}.

\textbf{Problem-solving.} \textsc{Problem-solving} involves identifying creative yet effective solutions by breaking from conventional approaches and overcoming cognitive biases such as functional fixedness, with the adaptability to generate innovative and practical outcomes. This domain of creativity requires both divergent and convergent thinking~\citep{chen2024boosting}.

\textbf{Research ideation.} \textsc{Research ideation} involves generating and evaluating novel scientific ideas by challenging existing paradigms, synthesizing concepts across disciplines~\citep{lu2024ai}. This creative process is crucial for driving breakthroughs that advance the discovery of new theories, methods, and applications capable of addressing complex global challenges.

\begin{table*}%[htbp]
\centering
\scalebox{0.95}{
\begin{tabular}{p{2cm}p{5cm}p{8cm}}
\toprule
    % \hline
    \textbf{Domain} & \textbf{Description} & \textbf{Sample task} \\
    % \hline
    \midrule
    \textsc{Creative writing} & Generating and identifying creative, engaging artistic material, such as movie plots or poems. & Given a movie synopsis \{synopsis\}, determine if its plot demonstrates originality and creative merit. \\
    % \hline
    \midrule
    \textsc{Problem-solving}
    & Finding creative solutions to unconventional problems. & Given items \{A, B, C\}, how can we achieve \{purpose\} when these items weren't designed for that purpose? \\
    % \hline
    \midrule
    \textsc{Research ideation} & Generating and identifying creative, novel scientific ideas in research writing. & Given a scientific research paper \{paper\}, determine whether its underlying ideas are sufficiently novel for it to be accepted. \\
    % \hline
    \bottomrule
\end{tabular}
}

\caption{\textbf{Creativity domain descriptions and sample tasks.}}
\label{table:domain_sample}
\end{table*}

\subsection{Datasets}
\label{sec:datasets}

To investigate the performance of existing creativity metrics across domains, we curate a dataset for each domain, with examples described in~\mytable{table:domain_sample}. Each dataset contains two sets of samples (creative and uncreative) labeled based on human perception of creativity.~\mysec{app:data_examples} presents representative creative and uncreative examples for each domain. To validate reliability, we conduct a pairwise verification task in which annotators identify the more creative example from 30 random pairs per dataset, achieving 73\% agreement with the human-labeled creative samples. Annotation was conducted by six graduate students with professional English proficiency (CEFR level $\ge$ B2). Each data point was independently annotated by two annotators, and inter-annotator agreement was subsequently computed on the paired annotations. Annotation instructions are provided in~\mysec{sec:human_annotation_ins}.

For \textsc{Creative writing}, we build upon datasets from \citet{tian2024a}, which showed LLM-generated narratives were less diverse in story arcs and had worse pacing compared to human narratives; we therefore label the dataset's 150 LLM-generated narratives as uncreative and its 150 human narratives as creative, with supporting details in~\mysec{app:movie_data_detail}. We augment the dataset with human-written synopses scraped from movie summary sites \href{https://moviepooper.com/}{MoviePooper} and \href{https://themoviespoiler.com/}{The Movie Spoiler}.

For \textsc{Problem-solving}, we adapt the MacGyver dataset for functional fixedness~\citep{tian2024b}, which contains a collection of problem-solving scenarios designed to assess the ability to think beyond conventional uses of everyday objects. We filter $573$ unconventional solutions and extract the tools mentioned in each problem to maintain similar settings across examples. We then augment the dataset by prompting LLMs to generate corresponding conventional problem-solution pairs, the validity of which we verify through human annotation. Both unconventional and conventional solutions were carefully rephrased to ensure consistent expression and style across the dataset.

For \textsc{Research ideation}, we collect data from OpenReview submissions to ICLR 2024, which has published soundness, contribution, and presentation scores for all accepted and rejected submissions. To control for variables other than creativity in research ideas, we group the papers according to their soundness and presentation scores.~\mysec{app:scientific_ideation_data} details the curation process, which enables us to focus on the creativity of papers as the primary differentiating factor between accepted and rejected papers as implemented by~\citet{zhao2025review}. We balance our sample size across groups, resulting in a final dataset of $513$ creative/accepted papers and $513$ uncreative/rejected papers.

\section{Creativity metrics selected for evaluation}

To comprehensively assess creativity across the three domains, we select four representative metrics that evaluate creativity from multiple levels of granularity: Creativity Index, Perplexity, syntactic templates, and LLM-as-a-Judge, as illustrated in~\myfig{fig:metrics_intro}. 
Together, these metrics provide a broad and multifaceted perspective on creativity—capturing variations at the token level, n-gram patterns, part-of-speech and syntactic structures, as well as semantic and holistic evaluation dimensions.

\textbf{Creativity Index (CI)} quantifies textual originality by measuring overlap with web corpora with the \textsc{DJ Search} algorithm~\citep{lu2025salieri}. It estimates $L$-uniqueness, the proportion of words appearing in novel $n$-gram contexts of length $L$ or greater, and averages these scores across a range of $L$ values. In this study, we focus on exact verbatim matches and implement CI using the Infini-Gram algorithm~\citep{liu2024infinigramscalingunboundedngram}. \mysec{app:metric_implementation} provides additional details and shows that near-verbatim matching yields results similar to exact matching.

\textbf{Perplexity (PPL)} measures the statistical unexpectedness of text based on probability distributions \citep{jelinek1977}.  It indicates how well a probability model predicts a sample, where lower probability is higher perplexity and therefore the content is less expected and possibly more creative.  

\textbf{Syntactic templates} examine structural diversity in text by identifying commonly-used part-of-speech patterns (\eg [DET NOUN VERB DET ADJ NOUN])~\citep{shaib2024}. It uses three key metrics:
\begin{itemize}
    \item \textbf{CR-POS:} Compression ratio of POS tag sequences; lower values indicate more varied syntax.
    \item \textbf{Template rate:} Fraction of texts containing at least one common template; lower values suggest greater structural originality.
    \item \textbf{Templates-per-token (TPT):} Number of templates normalized by text length; lower values denote more syntactic diversity.
\end{itemize}
While effective for detecting repetition, this metric captures structural rather than semantic creativity.

\textbf{LLM-as-a-Judge} uses LLMs to assess creativity at the conceptual level, complementing the linguistic metrics above. Building on recent work highlighting LLMs' promise for creativity assessment~\citep{organisciak, distefano2024, luchini2025automated, luchini2025multilingual}, we conduct extensive ablations across various dimensions, \eg prompt sensitivity, model bias, model consistency, and potential data contamination, as described in~\mysec{sec:llm_ablation}, and report the results of the best-performing configuration. Our approach employs a multi-pronged evaluation framework that targets idea-level creativity through chain-of-thought prompting, aspect-based evaluation, and clearly defined scoring criteria.

Our prompting strategies are tailored to each domain. For \textsc{Creative writing}, we evaluate background setup, logic, development, and originality on a 1–5 scale. For \textsc{Problem-solving}, we evaluate problem difficulty, solution unconventionality, and solution efficiency on a 1-5 scale. For \textsc{Research ideation}, we assess soundness (1-4), presentation (1-4), contribution (1-4), and overall quality (1-10) using domain-appropriate scales.

This approach captures conceptual creativity that might be missed by purely linguistic or statistical measures, providing a more holistic assessment framework. Detailed prompts for each domain are provided in the~\mysec{sec:llm_eval_prompts}.

\section{Results and analysis}

\subsection{Quantitative results}

\begin{table*}[t]
\centering
\scalebox{0.85}{
\begin{tabular}{llccccc}
\toprule
\multirow{2.5}{*}{\bf Domain}
& \multirow{2.5}{*}{\bf Truth label}
& \multirow{2.5}{*}{\bf CI $\uparrow$} 
& \multirow{2.5}{*}{\bf PPL $\uparrow$}
& \multicolumn{2}{c}{\bf Syntactic templates}
& \multirow{2.5}{*}{\bf LLM-as-a-Judge $\uparrow$} \\
\cmidrule(lr){5-6} 
& & & & CR-POS $\downarrow$ & $\geq$1 Template $\downarrow$\\
\midrule
% movie: creative writing
\multirow{2}{*}{\parbox{2cm}{\textsc{Creative writing}}}
    & human expert & \textbf{0.72} \ensuremath{\pm} 0.06 & 7.03 \ensuremath{\pm} 3.97 & \textbf{5.474} & \textbf{77.3}\% (\textbf{0.007}) & \textbf{2.69} \ensuremath{\pm} 0.76 \\
    & LLM-generated & 0.53 \ensuremath{\pm} 0.06 & \textbf{7.36} \ensuremath{\pm} 4.03 & 6.028 & 96.7\% (0.013) & 2.02 \ensuremath{\pm} 0.55 \\

% ff: problem solving
\midrule
\multirow{2}{*}{\parbox{2cm}{\textsc{Problem-solving}}}
    & unconventional & \textbf{0.76} \ensuremath{\pm} 0.08 & 13.98 \ensuremath{\pm} 4.65 & 6.158 & 38.7\% (0.039) & \textbf{3.02} \ensuremath{\pm} 0.87 \\
    & conventional & 0.75 \ensuremath{\pm} 0.09 & \textbf{14.11} \ensuremath{\pm} 4.85 & \textbf{6.152} & \textbf{31.9}\% (\textbf{0.033}) & 2.60 \ensuremath{\pm} 1.03 \\

% paper: research ideation
\midrule
\multirow{2}{*}{\parbox{2cm}{\textsc{Research ideation}}}
    & accepted & 0.71 \ensuremath{\pm} 0.05 & \textbf{8.62} \ensuremath{\pm} 1.62 &  5.684 & \textbf{54.6}\% (0.004) & 2.84 \ensuremath{\pm} 0.54 \\
    & rejected & 0.71 \ensuremath{\pm} 0.05 & 8.50 \ensuremath{\pm} 1.59 & \textbf{5.657} & 63.2\% (0.004) & \textbf{2.85} \ensuremath{\pm} 0.19 \\

\bottomrule
\end{tabular}
}
\caption{\textbf{Metric performance across creative domains.} The metrics exhibit inconsistent performance across domains: CI reflects linguistic diversity and is sensitive to the choice of L; PPL captures fluency rather than conceptual novelty; syntactic templates identify structure but not ideas; and LLM-as-a-Judge offers moderate performance with biased and unstable results.}
\label{table:all_metrics_results}
\end{table*}

We present the evaluation results of each metric across three creativity domains in~\mytable{table:all_metrics_results}. The results reveal that the metrics lack consistency, both within individual domains and across domains.

\textbf{Creativity Index (CI).}
CI effectively captures the differences in \textsc{Creative writing}; however, it yields comparable scores between unconventional and conventional solutions in \textsc{Problem-solving}, as well as between accepted and rejected paper introductions in \textsc{scientific ideation}. 
As discussed in~\mysec{unique_n_gram_analysis}, prior work often uses unique n-grams to evaluate novelty~\citep{crossley2016idea, merrill2024evaluating}. However, our results suggest CI reflects primarily lexical diversity which is more relevant in the writing domain rather than originality of ideas, limiting its effectiveness in capturing conceptual creativity. Furthermore,~\mysec{ci_L} highlights additional factors, such as the choice of the n-gram range (L uniqueness), which can significantly influence the CI scores and ultimately undermine the reliability of reflecting the linguistic creativity itself. Together with its dependence on the reference corpus, this strong sensitivity limits CI’s generality across domains and evaluation setups.

\textbf{Perplexity (PPL).}
From the PPL results, we observe that in \textsc{Creative writing} and \textsc{Problem-solving}, LLM-generated plots and conventional solutions tend to exhibit higher PPL, whereas in \textsc{Research ideation}, rejected paper introductions show lower PPL. This inconsistency highlights the inability of PPL to reliably distinguish creative from uncreative content.

Although some studies have proposed PPL as an indicator of surprise, diversity, or creativity~\citep{basu2021mirostat}, it was originally developed as a fluency metric and is heavily shaped by trade-offs between model confidence and output smoothness~\citep{guo2024curiousdeclinelinguisticdiversity}. Our findings show the score distributions for creative and uncreative texts overlap substantially across all domains, offering little discriminative signal. We therefore suggest PPL remains useful for monitoring fluency or guiding fine-tuning~\citep{guo2024curiousdeclinelinguisticdiversity, wu2025mitigating}, but should not be treated as a reliable metric of creativity.

\textbf{Syntactic templates.}
In \textsc{Problem-solving} and \textsc{Research ideation}, unconventional solutions and accepted papers tend to exhibit higher CR-POS scores and elevated template rates, including templates per token (TPT). In~\mysec{template_analysis}, we conduct a qualitative analysis of the templates used in \textsc{Creative writing}, revealing LLM-generated narratives often rely on repetitive and formulaic syntactic structures to introduce plot developments, in contrast to the more diverse constructions found in human-written content.
While syntactic templates are designed to evaluate linguistic creativity beyond token-level lexical variation, our results indicate this metric remains insufficient for capturing conceptual creativity—particularly in domains where novel ideas are often expressed using formulaic language. However, template-based analysis can still be useful for characterizing stylistic creativity and for identifying recurring structural patterns in text. Future adaptations could improve the value of syntactic templates by balancing the number of top templates selected with computational efficiency or combining template analysis with semantic measures of idea-level novelty.

\textbf{LLM-as-a-Judge.}
We conduct extensive ablations across four dimensions (prompt sensitivity, model bias, model consistency, and potential data contamination) in~\mysec{sec:llm_ablation} and report the best-performing results to evaluate the strongest configuration for LLM-as-a-Judge.
Based on our carefully designed evaluation prompts in~\mysec{sec:llm_eval_prompts}, LLMs demonstrate moderate performance. We tested multiple LLMs and reported the results from the best-performing model, Claude 3.7 Sonnet using default hyperparameters. However, a closer inspection reveals limitations in LLM-based evaluations, including the tendency toward one-sided predictions, the weak correlation with human perceptions, and instability under minor prompt variations. We report the details in~\mysec{sec:llm_eval_limit}.

\subsection{Analysis and discussion}
\label{sec:analysis}
\subsubsection{\texorpdfstring{Creativity Index: Idea originality $\neq$ unique n-grams in the content}{Creativity Index: Idea originality != unique n-grams in the content}}

\label{unique_n_gram_analysis}

\citet{crossley2016idea} explore the relationship between linguistic features and idea generation, operating under the assumption that "essays with greater originality would contain a greater number of unique n-grams," where originality is defined as "how different an individual’s ideas are from others’ ideas." However, our results, as reflected by CI in \textsc{Problem-solving} and \textsc{Research ideation}, show that texts containing original ideas do not necessarily exhibit a high frequency of unique n-grams.
On the other hand,~\citet{merrill2024evaluating} assess the novelty of LLM-generated versus human-written text through n-gram comparisons, where "novelty" is defined as the presence of n-grams not found in a reference corpus. While this approach aligns with our findings in \textsc{Creative writing}, where LLM-generated text contains fewer unique n-grams, it is important to note the presence of unique n-grams primarily indicates lexical rarity or uncommon phrasing, rather than genuine novelty or creativity at the conceptual or ideational level.

\subsubsection{Creativity Index: What affects the credibility of the Creativity Index?}
\label{ci_L}

\textbf{Data and corpora nature.} Although CI appears to perform well in the \textsc{Creative writing}, the dataset excludes well-known movie synopses from the human-written plots to avoid data leakage, which likely results in higher CI scores for human-written content. By design, CI quantifies creativity through n-gram comparisons between generated content and a reference corpus. Given that our less creative, human-annotated samples in \textsc{Creative writing} were generated by LLMs, it is possible the models were trained on corpora similar to the one used for comparison.

\begin{figure}[t]
\centering
  \includegraphics[width=\columnwidth]{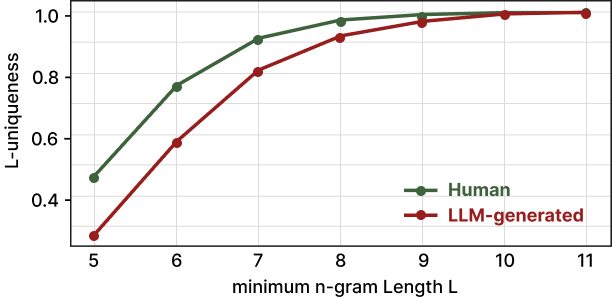}
  \caption{\textbf{L-uniqueness vs. minimum n-gram length L.} Creativity Index is sensitive to L.
        }
        \label{fig:L-uniqueness_min_n-gram}
\end{figure}

\begin{figure}[t]
\centering
  \includegraphics[width=\columnwidth]{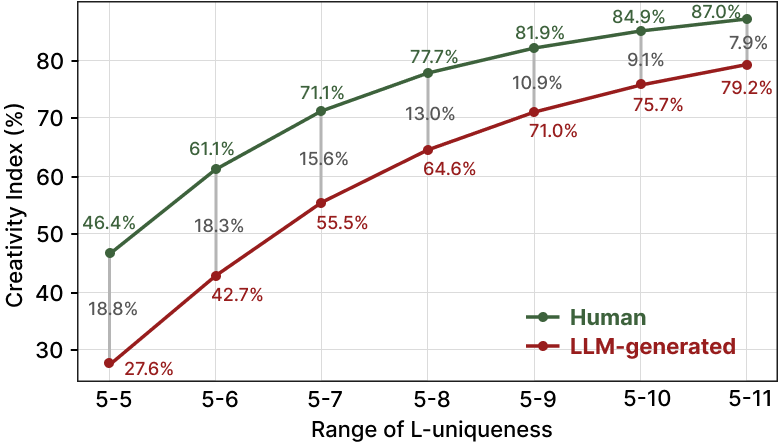}
  \caption{\textbf{CI value vs. L-uniqueness range.} As L-uniqueness increases, the CI difference decreases.
        }
        \label{fig:CI_value_L-uniqueness}
\end{figure}

\textbf{Range of L-uniqueness.} In the original study, the choice of L (the range of n-gram lengths) used for CI computation is condition-specific. As shown in~\myfig{fig:L-uniqueness_min_n-gram}, CI is particularly sensitive to the choice of L. When the range is expanded from 5–7 to 5–11, the CI score differences between human-written plots and LLM-generated content become negligible as shown in~\myfig{fig:CI_value_L-uniqueness}. \mytable{tab:corpus_domain_ranges} demonstrates that this sensitivity also persists across different reference corpora, specifically (1) Pile-val and (2) Dolma-v1.6-sample, where CI scores are computed under multiple L-ranges (5–5, 5–7, 5–9, and 5–11). This sensitivity limits CI's generalizability across different domains and content types and raises concerns about the discriminative power and potential bias in its ability to reflect true linguistic creativity.

\begin{table}[htbp]
\centering
\footnotesize
\setlength{\tabcolsep}{4pt}

\textbf{(1) Corpus: \texttt{Pile-val}}\\[2pt]
\begin{tabular}{p{1.5cm}p{2.3cm}cccc}
\toprule
\textbf{Domain} & \textbf{Truth label} & \textbf{5--5} & \textbf{5--7} & \textbf{5--9} & \textbf{5--11} \\
\midrule
\multirow{2}{*}{\parbox{2cm}{\textsc{Creative writing}}}  & human expert & 0.65 & 0.82 & 0.89 & 0.92 \\
                  & LLM-generated & 0.47 & 0.71 & 0.82 & 0.87 \\
\midrule
\multirow{2}{*}{\parbox{2cm}{\textsc{Problem-solving }}}  & unconventional & 0.69 & 0.86 & 0.91 & 0.94 \\
                  & conventional & 0.69 & 0.85 & 0.91 & 0.94 \\
\midrule
\multirow{2}{*}{\parbox{2cm}{\textsc{Research ideation}}} & accepted & 0.65 & 0.81 & 0.88 & 0.91 \\
                  & rejected & 0.64 & 0.80 & 0.87 & 0.91 \\
\bottomrule
\end{tabular}

\vspace{6pt}

\textbf{(2) Corpus: \texttt{Dolma-v1.6-sample}}\\[2pt]
\begin{tabular}{p{1.5cm}p{2.3cm}cccc}
\toprule
\textbf{Domain} & \textbf{Truth label} & \textbf{5--5} & \textbf{5--7} & \textbf{5--9} & \textbf{5--11} \\
\midrule
\multirow{2}{*}{\parbox{2cm}{\textsc{Creative writing}}}  & human expert & 0.48 & 0.72 & 0.82 & 0.87 \\
                  & LLM-generated & 0.26 & 0.53 & 0.69 & 0.78 \\
\midrule
\multirow{2}{*}{\parbox{2cm}{\textsc{Problem-solving }}}  & unconventional & 0.52 & 0.76 & 0.85 & 0.89 \\
                  & conventional & 0.52 & 0.75 & 0.85 & 0.89 \\
\midrule
\multirow{2}{*}{\parbox{2cm}{\textsc{Research ideation}}} & accepted & 0.50 & 0.71 & 0.81 & 0.86 \\
                  & rejected & 0.50 & 0.71 & 0.81 & 0.86 \\
\bottomrule
\end{tabular}

\caption{\textbf{Performance across corpora and domains.}
Scores increase with wider ranges, with substantial overlap between labels across domains.}
\label{tab:corpus_domain_ranges}
\end{table}

\subsubsection{Syntactic templates: Differences in LLM-generated vs. human-written content}
\label{template_analysis}
\begin{figure}[t]
\centering
  \includegraphics[width=\columnwidth]{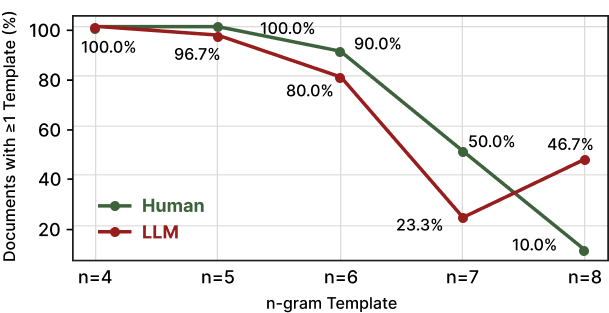}
  \caption{\textbf{Template rate by n-gram template.} As the top 100 most common templates are selected, their actual occurrence rate varies with n. Note that the sudden spike at n=8 is a methodological artifact due to the top-100 selection constraint.}
  \label{fig:template_rate_n-gram}
\end{figure}

While we would theoretically expect a gradual decrease in template rate as the n-gram size increases, we observe a sharp rise between the 7-gram and 8-gram (from 23.3\% to 46.7\%), as shown in~\myfig{fig:template_rate_n-gram}. This spike can be attributed to the methodology in the original study, which restricts analysis to the top 100 most frequently occurring templates for computational efficiency. As a result, LLM-generated content may produce a set of atypical 8-gram templates that are underrepresented in shorter n-gram ranges (n = 4–7).

\begin{figure}[t]
\centering
  \includegraphics[width=\columnwidth]{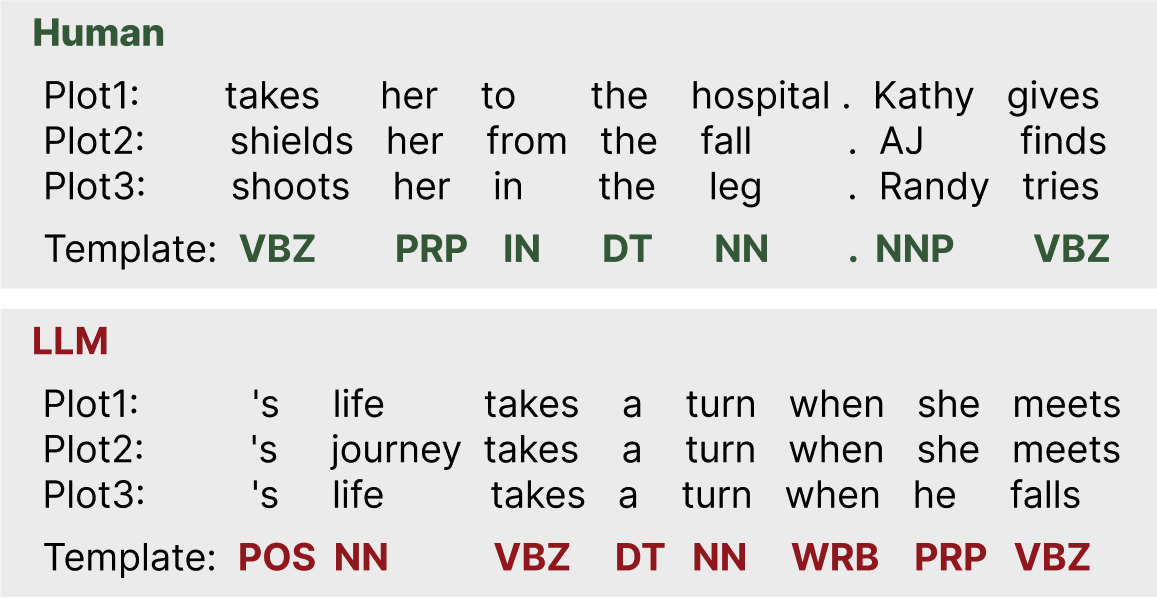}
  \caption{\textbf{Example of the 8-gram template.} LLM-generated plots reuse certain sentences, whereas human-written plots display more varied word choices.
        }
        \label{fig:example_8-gram_template}
\end{figure}

\myfig{fig:example_8-gram_template} reveals a recurring pattern in LLM-generated plots. For example, the sentence \textit{"[noun]'s life/journey takes a turn when [noun] [verb]..."} is frequently used to mark plot twists. In contrast, human-written texts demonstrate much greater syntactic variety within the same 8-gram constraints. When similar expressions appear in human-authored templates, they typically occur as part of shorter phrases embedded in more varied sentence structures,~\eg \textit{"The next day, [noun] [verb]...."}
The findings suggest the syntactic template metric can effectively capture recurring structural patterns. In \textsc{Creative writing}, our analysis highlights that LLMs may rely on specific structures to introduce narrative shifts, which potentially contributes to the perception of lower creativity in their storytelling.

\subsubsection{LLM-as-a-Judge's limitation in reflecting human-perceived creativity}
\label{sec:llm_eval_limit}
In~\mytable{table:all_metrics_results}, LLMs appear to demonstrate a moderate ability to distinguish the level of creativity in ideas across different domains. However, our initial experiments reveal LLMs struggle to reliably assess whether a given piece of content exhibits conceptual creativity when prompted directly. Even with the use of carefully designed prompts as detailed in~\mysec{sec:llm_eval_prompts}, LLM-based evaluations still display several notable limitations.

\begin{figure}[t]
\centering
  \includegraphics[width=0.95\columnwidth]{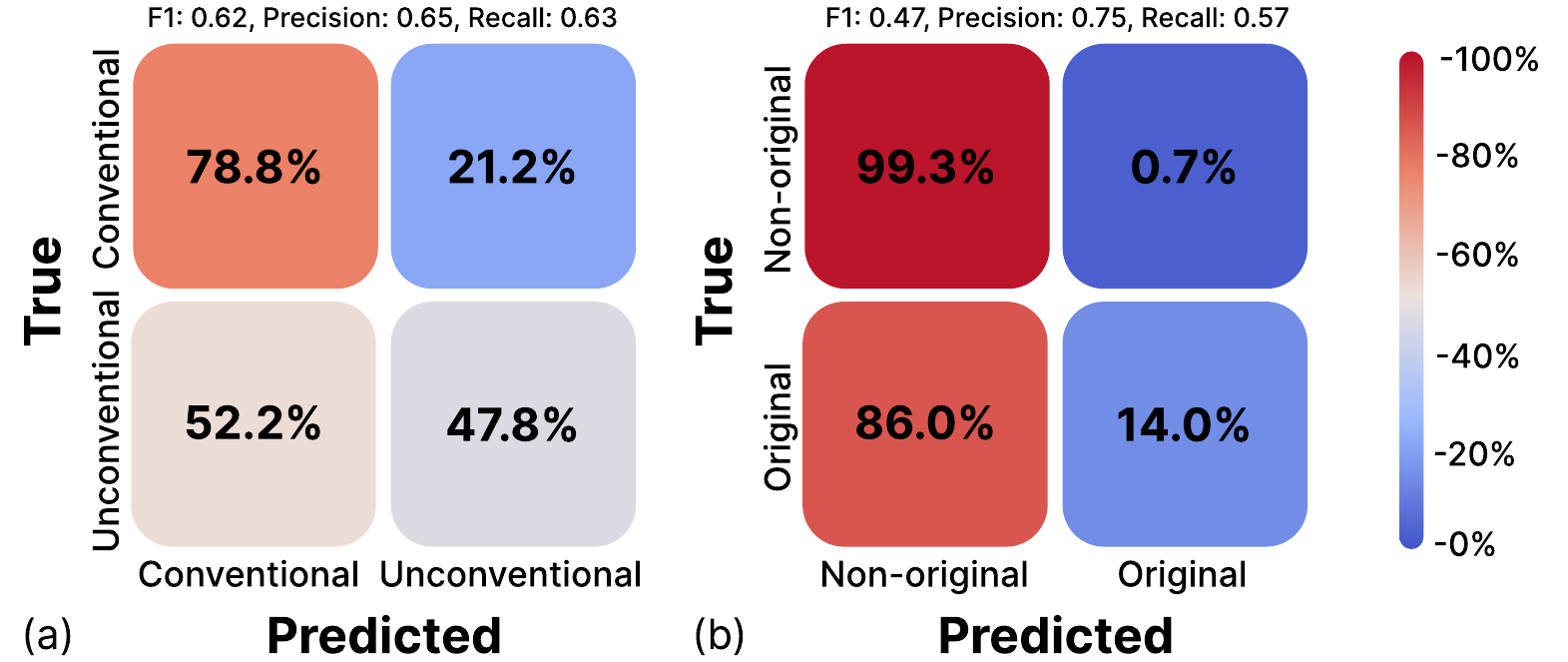}
      \caption{\textbf{LLM evaluation results with F1, precision, recall, and confusion-matrix reported.} (a) \textsc{Problem-solving}: LLMs tend to rate the solutions to be conventional; (b) \textsc{Creative writing}: LLMs tend to rate the plots to be non-original.}
  \label{fig:lm_eval_confusion_matrix}
\end{figure}

\textbf{Bias toward one-sided predictions.}
We observe that LLM's predictions tend to favor one side of the answers. Specifically, although the overall accuracy in \textsc{Problem-solving} appears reasonable,~\myfig{fig:lm_eval_confusion_matrix}(a) shows the model disproportionately classifies responses as “Unconventional.” Similarly,~\myfig{fig:lm_eval_confusion_matrix}(b) suggests that when applying a reasonable threshold of 4 out of 5 (according to the rubric) to distinguish original from non-original plots, our LLM judge tends to predict most content as non-original. Together with the F1 scores, accuracy, and confusion matrices, these results indicate the model's predictions are only marginally better than random guessing, highlighting a biased evaluation pattern that undermines reliability.

\textbf{Failure under joint novelty--acceptability filtering.} To further assess the ability of LLM-based judges to reflect human-perceived conceptual creativity, we jointly consider the dimensions of novelty and acceptability by explicitly incorporating multiple creativity-relevant criteria. Using multidimensional LLM evaluation prompts provided in~\mysec{sec:llm_eval_prompts}, we filter data points to retain only those for which both novelty-related and acceptability-related scores exceed predefined thresholds. The retained samples are treated as the \emph{predicted creative subset} and the excluded set as the \emph{predicted uncreative subset}. We then compare these predictions against human ground truth.

For \textsc{Creative writing}, we retain samples with LLM ratings of \texttt{Logic} $\geq 4$ (\textit{"Mostly logical and well-structured"}) and \texttt{Originality} $\geq 4$ (\textit{“Refreshingly different with clever elements”}). For \textsc{Research ideation}, we retain samples with \texttt{Presentation} $\geq 3$, \texttt{Soundness} $\geq 3$, and \texttt{Contribution} $\geq 3$. As shown in Table~\ref{tab:domain_metrics}, the resulting accuracies are 0.51 and 0.54 respectively—both close to random guessing. This indicates that even when jointly conditioning on multiple creativity-relevant dimensions, LLM-based judges do not robustly capture idea-level creativity.

\begin{table}[t]
    \centering
    \footnotesize
    \setlength{\tabcolsep}{3pt}
    \resizebox{\columnwidth}{!}{
    \begin{tabular}{p{1.5cm}cccccccc}
    \toprule
    \textbf{Domain} & \textbf{acc.} & \textbf{F1} & \textbf{prec.} & \textbf{rec.} & \textbf{TP} & \textbf{FP} & \textbf{FN} & \textbf{TN} \\
    \midrule
    \textsc{Creative writing}  & \multirow[c]{2}{*}{0.51} & \multirow[c]{2}{*}{0.04} & \multirow[c]{2}{*}{1.00} & \multirow[c]{2}{*}{0.02} & \multirow[c]{2}{*}{3}   & \multirow[c]{2}{*}{0}   & \multirow[c]{2}{*}{147} & \multirow[c]{2}{*}{150} \\
    \midrule
    \textsc{Research ideation} & \multirow[c]{2}{*}{0.54} & \multirow[c]{2}{*}{0.48} & \multirow[c]{2}{*}{0.55} & \multirow[c]{2}{*}{0.42} & \multirow[c]{2}{*}{218} & \multirow[c]{2}{*}{180} & \multirow[c]{2}{*}{295} & \multirow[c]{2}{*}{333} \\
    \bottomrule
    \end{tabular}
    }
    \caption{\textbf{Classification metrics by domain.}
    Accuracy (acc.), F1, precision (prec.), recall (rec.), and confusion-matrix counts are reported, with creative samples defined as the positive class (TP/FP/TN/FN).}
    \label{tab:domain_metrics}
\end{table}

\textbf{Weak correlation with human evaluators.} In \textsc{Research ideation}, we examine the “contribution” aspect of scientific paper introductions. The Pearson correlation coefficient between human reviewers’ scores and LLM evaluation scores is only 0.159. Similar weak correlations are observed across other dimensions (\eg presentation, soundness, and overall quality) and across different evaluation models. The highest observed correlation is still a weak association (0.234 for Contribution from GPT-4o), highlighting a significant mismatch between LLM judgments and human evaluators.

\textbf{Instability under minor prompt variations.} When the prompt in \textsc{Problem-solving} is slightly twisted from “\textit{...is the solution conventional?}” to “\textit{...unconventional?}”, the model alters its overall preferences. Moreover, approximately 20\% of the evaluations produced contradictory judgments for the same solution, depending solely on phrasing. This further illustrates the fragility and inconsistency of LLM-based evaluation in assessing creativity and originality.

\subsubsection{Improving LLM-as-a-Judge via chain-of-thought and rubric-based prompting}
\label{improve_llm_eval}

\textbf{Chain-of-thought}~\citep[CoT;][]{wei2022chain} assists LLMs in better adhering to evaluation instructions and consequently refining their final output scores. For example, in the third round of dialogue, the Claude agent revises (soundness, presentation, contribution, overall) evaluation scores from (3.5, 3, 3.5, 7) to (3, 3, 3, 6), with an explanation "\textit{...However, for a conference with only 30\% acceptance rate, I must be stringent. The paper, while valuable, doesn't appear to represent a breakthrough..., warranting a "Weak Accept" rating.}" This demonstrates the model’s ability to self-correct and adjust its assessment based on the provided context and constraints.

\textbf{Rubric-based prompting}
incorporates orthogonal dimensions into the evaluation, allowing LLMs to better isolate and focus on the assessment of originality. 
Additionally, a clear ranking criterion could help LLMs anchor more precise judgments, rather than assigning numerical scores arbitrarily. Our prompts are presented in~\mysec{sec:llm_eval_prompts}.

\subsubsection{Concept-level measurements: Semantic Novelty Score (SNS)}
To explore potential approaches for evaluating conceptual creativity, we design an exploratory experiment, a two-stage pipeline that produces a Semantic Novelty Score (SNS), to investigate whether concept-level semantic distances can serve as a proxy for conceptual creativity. Specifically, we first employ \texttt{gemini-2.5-flash} as a semantic parser to extract the five most distinct core concepts ($|C| = 5$) from each generated response. Second, we use the \texttt{text-embedding-004} high-dimensional embedding space as a continuous proxy for a semantic graph, mapping the extracted concepts into this space. We then calculate the pairwise semantic distance, defined as \texttt{$1 -$ cosine similarity}, across all unique concept pairs $P$. The SNS is defined as the average over these pairs:

\[
\mathrm{SNS} = \frac{1}{|P|} \sum_{(c_i, c_j) \in P} \left( 1 - \mathrm{sim}(v_i, v_j) \right).
\]

\mytable{tab:sns_by_domain} presents evaluation results on subsets of 30 creative and 30 uncreative samples from each of the three domains. The inability of the SNS to improve discrimination across domains underscores the challenges faced by automatic metrics in capturing idea-level novelty, even when semantic novelty is explicitly targeted.

\begin{table}[htbp]
    \centering
    \footnotesize
    \begin{tabular}{p{1.5cm}p{2.3cm}cc}
    \toprule
    \multirow{2.5}{*}{\bf Domain}
& \multirow{2.5}{*}{\bf Truth label}
& \multicolumn{2}{c}{\bf SNS}\\
\cmidrule(lr){3-4} 
& & Mean & std.\\
\midrule
    \multirow{2}{*}{\parbox{2cm}{\textsc{Creative writing}}}  & human expert & 0.447 & 0.030 \\
                      & LLM-generated & 0.421 & 0.084 \\
    \multirow{2}{*}{\parbox{2cm}{\textsc{Problem-solving}}}    & unconventional & 0.402 & 0.036 \\
                      & conventional & 0.402 & 0.035 \\
    \multirow{2}{*}{\parbox{2cm}{\textsc{Research ideation}}} & accepted & 0.420 & 0.049 \\
                      & rejected & 0.435 & 0.037 \\
    \bottomrule
    \end{tabular}
    \caption{\textbf{Semantic Novelty Score (SNS) by domain and label.}
    SNS does not consistently distinguish between creative and uncreative outputs across domains.}
    \label{tab:sns_by_domain}
\end{table}
\section{Conclusion}
Our analyses highlight the significant inconsistencies across various creativity metrics, demonstrating that each metric captures different dimensions of creativity. The performance of these metrics, particularly perplexity and syntactic templates, is often task-specific and not generalizable across domains. Furthermore, metrics like the Creativity Index and LLM-as-a-Judge have limitations in capturing deeper aspects of creativity, such as novelty and conceptual innovation. These findings underscore the necessity for developing more reliable and robust evaluation frameworks. Ultimately, advancing creativity assessment methods is essential for closing the gap between human and machine creative performance, guiding future research in this area.

\section*{Limitations}
The key contributions of our work include systematically demonstrating the limitations of existing metrics in capturing conceptual creativity, and highlighting inconsistencies across metrics and domains. While we recognize that developing more comprehensive and effective evaluation methods is a key direction for advancing creativity research, such methods are not proposed in this work.

Our study provides essential diagnostic groundwork for the field. Comprehensive, multi-domain analyses are indispensable precursors to robust solutions. By systematically revealing where and how current metrics fail—particularly their tendency to emphasize stylistic novelty while neglecting conceptual novelty—our findings provide actionable insight for both practitioners and researchers. Practitioners can interpret existing metrics with appropriate caution, and researchers can target the specific conceptual gaps that must be addressed in future work.

Future research can build on our findings by collecting high-quality, orthogonal datasets in underexplored domains such as musical composition, humor, or product design. These datasets would complement our current benchmarks and support the development of generalizable, multi-domain, conceptually aware creativity judges that capture signals beyond surface-level patterns. Once these datasets are available, we can fine-tune LLMs on conceptual rather than syntactical signals. We suggest the next step may be incorporating recent advances such as \citet{barrault2024large}’s Large Concept Model (LCM), which uses sentence-level embeddings as a proxy for concept-level semantics, and continuous chain‐of‐thought architectures~\citep{hao2024training}, which enable extended reasoning over idea chains. 
We believe that explainability could be a linchpin for creativity metrics that align more closely with human expectations and serve as a crucial component of future creativity evaluators.
\section*{Acknowledgments}
The authors thank the PlusLab members at UCLA and the anonymous reviewers for
their valuable feedback and helpful discussions. 
This work was supported in part by the National Science Foundation CAREER award \#2339766, an Amazon AGI foundation research award, and the National Science and Technology Council, Taiwan, under the Grant 114-2628-E-002-021-, and the Taiwan Centers of Excellence. Shao-Hua Sun was supported by the Yushan Fellow Program of the Ministry of Education, Taiwan.

\bibliography{custom}

\clearpage
\appendix
\section*{Appendix}

\begingroup
\hypersetup{colorlinks=false, linkcolor=black}
\hypersetup{pdfborder={0 0 0}}
\part{} % Start the appendix part
\parttoc % Insert the appendix TOC
\endgroup

\section{Data curation for scientific ideation}
\label{app:scientific_ideation_data}
We collect the paper from ICLR 2024 using the \href{https://openreview.net/}{OpenReview} and \href{https://arxiv.org/}{arXiv} APIs, and design a multi-step filtering process to reduce the impact of confounding factors when analyzing accepted and rejected papers. 

Alongside the accept or reject status, we collect the presentation, soundness, contribution, and overall scores for each paper. Scientific contribution is commonly used to represent creativity in the context of research ideation~\citep{zhao2025review}. To keep the focus on the research framing and ideation components, we filter for the Introduction section of papers, select papers based on the average of their overall review scores, and utilize soundness and presentation scores as control variables to reduce the influence of other factors.

ICLR 2024 received 7262 submissions, with 2185 accepted. We select the top and bottom 50\% of papers based on their average review scores, and then take the minimum number of papers between the two classes, resulting in 1093 accepted and 1093 rejected papers.

To further control for confounding review dimensions, we group papers by matching soundness and presentation scores. Each paper’s average score in these two dimensions is binned into four ranges (e.g., 2.0-2.4, 2.5-2.9, 3.0-3.4, 3.5-3.9), and papers are assigned to a group based on their bin combination. Within each group, we keep the contribution score as the varying factor and sampled equal numbers of accepted and rejected papers based on the smaller class size. This results in 16 balanced groups with 513 accepted and 513 rejected papers (N = 1026). This helps ensure that differences in contribution scores are not confounded by differences in other review criteria, allowing our dataset to focus on contribution, i.e., scientific creativity, in research ideation while reducing the influence of other review criteria.

\section{CI implementation}
\label{app:metric_implementation}
% \todo{Add details of near-verbatim match from rebuttal} 
For the results in~\mytable{table:all_metrics_results}, we compute CI with $L$-uniqueness in the range $5 \leq L \leq 7$, following Appendix B.3 of \citet{lu2025salieri}. We observe the same saturation behavior for $L \geq 7$ (values $\approx$ 1), and therefore adopt the same setting. We also consider the near-verbatim match method using Word Mover’s Distance. However, this approach has computational complexity $O(|d|^2|w|)$ for each document $d$ and n-gram $w$, making it impractical for our large, diverse datasets. Moreover, \citep{lu2025salieri} reported similar human-vs-LLM gaps whether using verbatim or verbatim+semantic matching. Since parts of the original infrastructure (e.g., Elasticsearch WIMBD indices) of the “near-match CI” concept in \citet{lu2025salieri} have been deprecated, we implemented a version using \texttt{Llama-3 8B Instruct} embeddings. Specifically, we apply text embeddings to the same index used for CI in our main experiments (\texttt{v4\_dolmasample\_olmo}), and evaluate this variant on balanced subsets of 30 creative and 30 uncreative samples from each of the three domains. The results in~\mytable{tab:domain_label_performance} closely mirror our CI findings: the creative and uncreative distributions overlap substantially, indicating that even the embedding-based evaluation is not discriminative as a metric for creativity. In addition, we observe the same sensitivity to the L-range, with CI scores increasing as
$L$ grows, as reported in~\mysec{ci_L}.

\begin{table*}[t]
\centering
\scalebox{0.9}{
\begin{tabular}{l l c c c c c}
\toprule
\textbf{Domain} & \textbf{Truth label} & \textbf{L5} & \textbf{L6} & \textbf{L7} & \textbf{L5--6} & \textbf{L5--7} \\
\midrule
    \multirow{2}{*}{\parbox{2cm}{\textsc{Creative Writing}}}  & human expert & 0.896 & 0.929 & 0.962 & 0.913 & 0.929 \\
                  & LLM-generated & 0.878 & 0.921 & 0.963 & 0.900 & 0.921 \\
    \multirow{2}{*}{\parbox{2cm}{\textsc{Problem Solving}}}   & unconventional & 0.900 & 0.957 & 0.981 & 0.928 & 0.946 \\
                  & conventional & 0.899 & 0.958 & 0.984 & 0.928 & 0.947 \\
    \multirow{2}{*}{\parbox{2cm}{\textsc{Research Ideation}}} & accepted & 0.897 & 0.939 & 0.973 & 0.918 & 0.936 \\
                  & rejected & 0.905 & 0.948 & 0.976 & 0.926 & 0.943 \\
\bottomrule
\end{tabular}
}
\caption{\textbf{Embedding-based CI performance by domain and truth label.} Scores increase with larger $L$ but overlap substantially between labels across domains.}
\label{tab:domain_label_performance}
\end{table*}

\section{Label verification for creative writing}
\label{app:movie_data_detail}

In the dataset from prior work~\citep{tian2024a}, turning points (markers of a story's pacing) were shown to have highly significant differences at tp$_4$ (mean\_diff = $-3.59$, $p = 8.7 \times 10^{-8}$) and tp$_5$ (mean\_diff = $-4.19$, $p = 5.4 \times 10^{-8}$), indicating that LLM narratives reach these late-stage beats much earlier than human ones. Similarly, the distribution of story arcs (general plotlines such as ``Man in Hole'' or ``Rags to Riches'') differs dramatically ($p = 1.9 \times 10^{-11}$). To further support our assumption, we have run some additional pairwise human annotations.

We take a random sample of 100 pairs of LLM and human narratives in our creative writing dataset, then recruit 18 well-trained independent annotators, with each pair annotated by at least two annotators. Annotators are shown two narratives, A and B, and then indicate which narrative they felt is more creative. We take the majority vote when annotators disagree. This allows us to compute how often a given LLM narrative will outperform its associated human narrative in creativity. Our analysis of the pairwise distribution reveals the following:

\begin{itemize}
    \item 88\% of the time (88/100 pairs), annotators choose a human narrative over an LLM one
    \item 12\% of the time (12/100 pairs), annotators choose an LLM narrative over a human one
    \item The 95\% confidence interval of LLM narratives being preferred over human narratives is [6.8\%, 17.2\%]
\end{itemize}
These findings show that only a small fraction of LLM narratives outperform their human counterparts, confirming that our coarse binary labels still capture the predominant trend.

\section{Human annotation instructions}
\label{sec:human_annotation_ins}
We present samples from each domain together with detailed instructions for the human annotation tasks. 

\textsc{Creative writing}:
\textit{Below are some movie plots. Select the one that is more creative. By defining creative, consider: How predictable is the plot? How original are the events and developments?}

\textsc{Problem-solving}:
\textit{"Below are some problem-solving question-answer pairs. Select the pair that is more unconventional."}

\textsc{Research ideation}:
\textit{"Below are some paper introductions. Select the one with higher contribution, bringing more new scientific breakthroughs and innovation to the research community."}

\section{Metrics yield misaligned performance across the same concept}
\label{movie_human}
We evaluate the metrics on three different versions of the same movie plots, which we curated to assess how well each metric handles variations in expression while preserving the underlying concept.

As shown in~\mytable{table:3_sources_movie_results}, all four metrics exhibit inconsistencies when evaluating semantically equivalent content. This finding further demonstrates that existing methods are not robust to distributional shifts in expression, highlighting their limitations in evaluating conceptual creativity across diverse surface forms.

\begin{table*}[htbp]
\centering
\begin{center}
\scalebox{0.9}{
    \begin{tabular}{llcccccccc}
    \toprule
    \multirow{2.5}{*}{\bf Source}
    & & \multicolumn{2}{c}{\bf CI$\uparrow$} 
    & \multicolumn{2}{c}{\bf PPL$\uparrow$} 
    & \multicolumn{2}{c}{\bf Syntactic templates$\downarrow$}
    & \multicolumn{2}{c}{\bf LLM-as-a-Judge$\uparrow$} \\
    \cmidrule(lr){3-4} \cmidrule(lr){5-6} \cmidrule(lr){7-8} \cmidrule(lr){9-10}
    && Mean & std. & Mean & std. & CR-POS$\downarrow$ & $\geq$1 Template (\%)$\downarrow$ & Mean & std.\\

    \midrule
    Wikipedia & & \textbf{0.72} & 0.05 & 10.09 & 3.33 & 5.085 & 70.0 (0.013) & \textbf{2.85} & 0.79 \\
    Spoiler   & & 0.66 & 0.04 & \textbf{10.87} & 3.34  & 5.437 & 85.0 (\textbf{0.012}) & \textbf{2.85} & 0.85 \\
    Pooper    & & 0.65 & 0.08 & 9.56 & 2.00  & \textbf{4.808} & \textbf{5.00} (0.018) & 2.65 & 0.85 \\

    \bottomrule
    \end{tabular}
}
\caption{\textbf{Metrics performance across three sources of movie plots.}
}
\label{table:3_sources_movie_results}
\end{center}
\end{table*}

\section{LLM-as-a-Judge ablation studies}
\label{sec:llm_ablation}
We conduct ablation studies on four dimensions: prompt sensitivity, model bias, model consistency, and potential data contamination in order to evaluate the strongest configuration of LLM-as-a-Judge as a creativity assessor.

\subsection{Prompt selection}
We test GPT-4o and Claude 3.7 Sonnet across domains with strategies including direct prompt, few-shot, score-based criteria, roleplaying with constraints, multi-aspect rubric-based prompt, CoT, and a combination of multi-aspect rubric with CoT, which yields the best performance and is therefore adopted, as shown in~\mysec{sec:llm_eval_prompts}.

\textbf{Direct prompts}: 
LLM judges provide baseline evaluations but often lack depth or discriminative power.

\textbf{Few-shot}:
Supplying 1-2 creative and uncreative examples does not improve performance, suggesting a limited ability to generalize in evaluating conceptual creativity.

\textbf{Score-based criteria on novelty}:
While criteria can help LLMs quantify evaluation, scores are often concentrated in the mid-to-high range (e.g., on a 1-5 scale, values of 1 or 2 are rarely used). In some cases, the model assigns nearly identical scores across all samples, indicating that criteria alone are insufficient.

\textbf{Roleplaying with constraints vs. multi-aspect rubric-based prompt}:
Simply instructing the model does not result in more discriminative judgments. For example, in scientific ideation, contribution scores under constraints are indistinguishable from those produced with score-based novelty prompts. In contrast, multi-aspect rubric with orthogonal dimensions better isolates confounding factors such as presentation and soundness, and provides judgments more specifically targeted at scientific creativity, as discussed in~\mysec{improve_llm_eval}.

\textbf{CoT}:
We notice performance improvement when we implement CoT by first asking LLM to provide analysis as to strengths and weaknesses, then asking LLM to score based on its answer. We also observe adjustments in evaluation patterns, as explained in~\mysec{improve_llm_eval}.

\subsection{Model bias}
We ablate evaluation models (Claude 3.7 Sonnet, GPT-o1, GPT-o3 mini, GPT-4o) with the final prompt on 60 creative writing examples: the accuracies(acc.) are 0.57, 0.53, 0.55, and 0.52, respectively, close to random guessing.~\mytable{tab:model_bias} presents F1, precision(prec.), and recall(rec.) scores, where the consistently low recall indicates that the models exhibit bias toward one-sided judgments, resulting in unstable and unreliable evaluations of creativity.

\begin{table}[htbp]
    \centering
    \begin{tabular}{p{2.8cm}p{0.6cm}p{0.6cm}p{0.6cm}p{0.6cm}}
    \toprule
    \textbf{Evaluate model} & \textbf{acc.} & \textbf{F1} & \textbf{prec.} & \textbf{rec.} \\
    \hline
    GPT-o1            & 0.53 & 0.38 & 0.67 & 0.36 \\
    GPT-4o            & 0.52 & 0.36 & 0.67 & 0.34 \\
    GPT-o3-mini       & 0.55 & 0.39 & 0.67 & 0.37 \\
    Claude 3.7 Sonnet & 0.57 & 0.47 & 0.77 & 0.57 \\
    \bottomrule
    \end{tabular}
    \caption{\textbf{Performance of evaluation models.} All models achieve near-random accuracy, with consistently low recall(rec.) values indicating a systematic bias toward one-sided judgments.}
    \label{tab:model_bias}
\end{table}

\subsection{Model consistency}
In addition to the prompt-twisting experiments in~\mysec{sec:llm_eval_limit}, we re-evaluate 20 problem-solving cases three times with Claude 3.7 Sonnet and measure the degree of agreement across runs. Only 40\% of cases were consistent across three runs, and over half of these were consistently incorrect, meaning that the LLM-as-a-Judge produced correct and consistent judgments for only about 15\% of the problem-solution pairs.

\subsection{Potential data contamination}\

We evaluate models with different knowledge cutoff dates on 74 scientific ideation cases across 16 control groups (see~\mysec{app:scientific_ideation_data}) to inspect the potential data contamination. Human review data show that in 14 of 16 groups, accepted papers received higher contribution scores than rejected ones, validating our controls. We consider Claude 3 Haiku (Aug 2023), GPT-o1/4o/o3-mini (Oct 2023), and Claude 3.7 Sonnet (Oct 2024). Since these cutoffs respectively predate, align with, and fall well after the ICLR 2024 deadline, this setup enables us to assess potential contamination. We then measure how often each model assigned higher contribution scores to accepted papers within each group.
% Define custom colors once, so you can adjust them later
\definecolor{DarkRed}{RGB}{180,0,0}
\definecolor{DarkGreen}{RGB}{0,120,0}

% Column types
\newcolumntype{G}{>{\centering\arraybackslash}p{1.2cm}} % thinner group col
\newcolumntype{C}{>{\centering\arraybackslash}p{1.7cm}} % all others

\begin{table*}[htbp]
\centering
\renewcommand{\arraystretch}{1.2}

\scalebox{0.95}{\begin{tabular}{GCCCCCCC}
\toprule
\textbf{Group} & \textbf{Truth label} & \textbf{Human scores} & \textbf{Claude 3 Haiku} & \textbf{GPT-o1} & \textbf{GPT-4o} & \textbf{GPT-o3-mini} & \textbf{Claude Sonnet 3.7} \\
\hline
\multirow{2}{*}{0} & accepted & \textcolor{DarkGreen}{2.33} & 4.00 & 3.00 & 3.50 & 3.00 & \textcolor{DarkGreen}{3.17} \\
                   & rejected & \textcolor{DarkGreen}{1.83} & 4.00 & 3.00 & 3.50 & 3.00 & \textcolor{DarkGreen}{3.00} \\
\hline
\multirow{2}{*}{1} & accepted & \textcolor{DarkRed}{2.30} & 4.00 & 3.33 & \textcolor{DarkRed}{3.00} & 3.00 & 2.83 \\
                   & rejected & \textcolor{DarkRed}{2.37} & 4.00 & 3.33 & \textcolor{DarkRed}{3.33} & 3.00 & 2.83 \\
\hline
\multirow{2}{*}{2} & accepted & \textcolor{DarkGreen}{2.30} & 4.00 & 3.00 & 3.00 & 3.00 & \textcolor{DarkGreen}{2.83} \\
                   & rejected & \textcolor{DarkGreen}{1.97} & 4.00 & 3.00 & 3.00 & 3.00 & \textcolor{DarkGreen}{2.33} \\
\hline
\multirow{2}{*}{4} & accepted & \textcolor{DarkGreen}{2.83} & 4.00 & \textcolor{DarkGreen}{3.33} & \textcolor{DarkGreen}{3.33} & \textcolor{DarkGreen}{3.67} & \textcolor{DarkGreen}{3.17} \\
                   & rejected & \textcolor{DarkGreen}{2.53} & 4.00 & \textcolor{DarkGreen}{3.00} & \textcolor{DarkGreen}{3.00} & \textcolor{DarkGreen}{3.33} & \textcolor{DarkGreen}{2.83} \\
\hline
\multirow{2}{*}{5} & accepted & \textcolor{DarkGreen}{2.87} & 4.00 & 3.00 & \textcolor{DarkRed}{3.00} & 3.00 & \textcolor{DarkGreen}{2.83} \\
                   & rejected & \textcolor{DarkGreen}{2.03} & 4.00 & 3.00 & \textcolor{DarkRed}{3.33} & 3.00 & \textcolor{DarkGreen}{2.67} \\
\hline
\multirow{2}{*}{6} & accepted & \textcolor{DarkGreen}{2.67} & 4.00 & 3.00 & 3.00 & 3.00 & \textcolor{DarkGreen}{2.83} \\
                   & rejected & \textcolor{DarkGreen}{2.40} & 4.00 & 3.00 & 3.00 & 3.00 & \textcolor{DarkGreen}{2.67} \\
\hline
\multirow{2}{*}{7} & accepted & \textcolor{DarkGreen}{2.75} & 4.00 & \textcolor{DarkGreen}{3.50} & \textcolor{DarkRed}{3.00} & 3.00 & \textcolor{DarkGreen}{3.25} \\
                   & rejected & \textcolor{DarkGreen}{2.55} & 4.00 & \textcolor{DarkGreen}{3.00} & \textcolor{DarkRed}{3.25} & 3.00 & \textcolor{DarkGreen}{2.25} \\
\hline
\multirow{2}{*}{8} & accepted & \textcolor{DarkRed}{2.77} & 4.00 & 3.00 & 3.00 & \textcolor{DarkGreen}{3.33} & 2.83 \\
                   & rejected & \textcolor{DarkRed}{2.80} & 4.00 & 3.00 & 3.00 & \textcolor{DarkGreen}{3.00} & 2.83 \\
\hline
\multirow{2}{*}{9} & accepted & \textcolor{DarkGreen}{2.87} & 4.00 & 3.33 & \textcolor{DarkGreen}{3.17} & 3.00 & \textcolor{DarkGreen}{3.00} \\
                   & rejected & \textcolor{DarkGreen}{2.57} & 4.00 & 3.33 & \textcolor{DarkGreen}{3.00} & 3.00 & \textcolor{DarkGreen}{2.83} \\
\hline
\multirow{2}{*}{10} & accepted & \textcolor{DarkGreen}{3.07} & 4.00 & \textcolor{DarkGreen}{3.33} & 3.17 & 3.00 & \textcolor{DarkGreen}{3.50} \\
                    & rejected & \textcolor{DarkGreen}{2.50} & 4.00 & \textcolor{DarkGreen}{3.00} & 3.17 & 3.00 & \textcolor{DarkGreen}{2.83} \\
\hline
\multirow{2}{*}{11} & accepted & \textcolor{DarkGreen}{2.93} & 4.00 & 3.00 & \textcolor{DarkRed}{3.17} & 3.00 & \textcolor{DarkGreen}{2.83} \\
                    & rejected & \textcolor{DarkGreen}{2.67} & 4.00 & 3.00 & \textcolor{DarkRed}{3.33} & 3.00 & \textcolor{DarkGreen}{2.50} \\
\hline
\multirow{2}{*}{13} & accepted & \textcolor{DarkGreen}{3.00} & \textcolor{DarkRed}{3.50} & \textcolor{DarkRed}{2.00} & \textcolor{DarkRed}{2.50} & \textcolor{DarkRed}{2.00} & \textcolor{DarkRed}{1.00} \\
                    & rejected & \textcolor{DarkGreen}{2.80} & \textcolor{DarkRed}{4.00} & \textcolor{DarkRed}{3.00} & \textcolor{DarkRed}{3.00} & \textcolor{DarkRed}{3.00} & \textcolor{DarkRed}{3.00} \\
\hline
\multirow{2}{*}{14} & accepted & \textcolor{DarkGreen}{3.07} & 4.00 & 3.33 & \textcolor{DarkRed}{3.17} & \textcolor{DarkRed}{3.33} & 2.83 \\
                    & rejected & \textcolor{DarkGreen}{2.77} & 4.00 & 3.33 & \textcolor{DarkRed}{3.50} & \textcolor{DarkRed}{3.50} & 2.83 \\
\hline
\multirow{2}{*}{15} & accepted & \textcolor{DarkGreen}{3.80} & 4.00 & 3.00 & 3.00 & 3.00 & 3.00 \\
                    & rejected & \textcolor{DarkGreen}{2.80} & 4.00 & 3.00 & 3.00 & 3.00 & 3.00 \\
\bottomrule
\end{tabular}}

\caption{\textbf{Performance of LLM judges with different cutoffs.} Comparison of human and models' evaluation across groups (\textcolor{DarkRed}{red = groups misaligned with truth labels}; \textcolor{DarkGreen}{green = groups aligned with truth labels}). Although Claude Sonnet 3.7 achieves the strongest alignment, this may be due to potential training data contamination rather than a genuine ability to evaluate conceptual novelty.}
\label{table:llm_contanmination}
\end{table*}

\mytable{table:llm_contanmination} shows that Claude 3 Haiku fails to provide informative evaluations. Among models with similar knowledge cutoffs, GPT-4o slightly outperforms o1 and o3 mini, suggesting that complex confounding factors make LLM-as-a-Judge unstable on this task. Finally, although Claude 3.7 Sonnet achieves the strongest alignment, this may be due to potential training data contamination rather than a genuine ability to evaluate conceptual novelty, raising further concerns about the reliability of LLMs as evaluators.

\section{LLM evaluation prompts}
\label{sec:llm_eval_prompts}
The prompts we used for LLM-as-a-Judge for each task are shown below.

\begin{myblock}{Domain - \textsc{Creative writing}}
'''Given a movie plot: ```\{MOVIE PLOT\}``` Analyze the plot in terms of the following four aspects: Background Setup, Logic, Development, and Originality. For each aspect, consider the following questions: Background Setup: How common or unique is the background setting? Logic: Is the timeline clear and are the cause-and-effect relationships logical?
Development: Are the twists well-developed? Is the buildup to the climax sufficient and engaging?
Originality: How predictable is the plot? How original are the events and developments? For each aspect, discuss the strengths and weaknesses of the plot. Provide a thoughtful and critical analysis.''',

'''Given the movie plot and your analyzation across four aspects, which The criteria for each aspect are as follows:

A. Background Setup
1. Very common or clichéd background, lacks uniqueness.
2. Somewhat common, but with a few unique elements.
3. Generally unique, with some familiar elements.
4. Mostly unique background and setup.
5. Exceptionally unique background setup.

B. Logic
1: Poor logic, many inconsistencies or plot holes.
2: Somewhat confusing or unrealistic transitions.
3: Generally coherent with minor issues.
4: Mostly logical and well-structured.
5: Strong internal logic and smooth cause-effect flow.

C. Development
1: Underdeveloped or abrupt changes.
2: Weak buildup, limited depth.
3: Average development, a few effective moments.
4: Good buildup with engaging twists.
5: Excellent pacing, buildup, and well-executed climax.

D. Originality
1: The plot is highly predictable or derivative.
2: Mostly familiar with few surprises.
3: Balanced mix of familiar and fresh ideas.
4: Refreshingly different with clever elements
5: Exceptionally creative and surprising or inspiring throughout.

Ensure that each aspect is orthogonal to the others—for example, the uniqueness of background setup should not affect the evaluation of plot originality and development. 
You are a rigorous reviewer. Provide the scores in the following format: [[Background Setup: X]], [[Logic: X]], [[Development: X]], [[Originality: X]].'''
\end{myblock}

\begin{myblock}{Domain - \textsc{Problem-solving}}
'''Given the problem: ```\{PROBLEM\}``` and the solution: ```\{SOLUTION\}```, analyze the following three aspects, taking into account the corresponding dimensions:
1. Problem Difficulty: Is the problem conceptually complex, or does it involve multiple steps or layers of reasoning? Are there hidden constraints or non-obvious elements that make the problem harder?
2. Unconventionality of the Solution: Does the solution follow a standard method, or does it use a surprising or creative strategy? Would this solution surprise someone familiar with typical methods for this problem? Are the tools, concepts, or constraints used in unexpected ways — beyond their typical or intended function? 
3. Efficiency of the Solution: Does it reach the goal using the fewest steps or resources possible? Could the solution be improved or simplified without losing correctness? Carefully summarize and provide a thorough assessment of each aspect.''',

'''Given the problem, solution, and your analyzation across three aspects, provide a score for each aspect accordingly, which The criteria for each aspect are as follows:
A. Problem Difficulty  
1: Very Easy (trivial or commonly seen problems; requires little reasoning)
2: Easy (straightforward with minimal abstraction or complexity)
3: Moderate (requires some thought or mild creativity; manageable)
4: Challenging (requires significant thought, creative insight, or multiple reasoning steps)
5: Very Challenging (highly complex or abstract; involves deep insight or several layers of reasoning)

B. Unconventionality
1: Very Conventional (the most expected approach; standard method anyone would think of first)
2: Conventional (follows a known method that is not surprising, but slightly less obvious)  
3: Moderately Unconventional (mix of familiar and unexpected methods or framings)
4: Unconventional (surprising to an expert; includes a creative or less typical approach)  
5: Highly Unconventional (reframes the problem or uses tools and constraints in a deeply creative, unexpected way)

C. Efficiency
1: Very Inefficient (excessive or redundant steps; far from optimal)  
2: Inefficient (works, but could clearly be improved with fewer steps or resources)
3: Moderately Efficient (reasonable efficiency; could still be optimized)
4: Efficient (solid balance of steps and resources; minimal waste)
5: Very Efficient (achieves the goal in the fewest possible steps or resources with no clear room for improvement)

Ensure that each aspect is orthogonal to the others—for example, Unconventionality should not be affected by the Efficiency or the Problem Difficulty. Strictly follow the criteria and provide the scores in the following format: [[Problem Difficulty: X]], [[Unconventionality: X]], [[Efficiency: X]].'''
\end{myblock}

\begin{myblock}{Domain - \textsc{Research ideation}}
''' You are a reviewer for a top conference, the acceptance rate for this conference is only around 20-30\%. Given the introduction of the paper: ```\{PAPER CONTENT\}```, carefully summarize and provide a thorough assessment of the strengths and weaknesses of the paper, touching on each of the following dimensions: 1. Originality: Are the tasks or methods new? Is the work a novel combination of well-known techniques?  Is it clear how this work differs from previous contributions? 2. Quality: Is the submission technically sound? Are claims well supported (e.g., by theoretical analysis or experimental results)? 
Are the methods used appropriate? Is this a complete piece of work or work in progress? Are the authors careful and honest about evaluating both the strengths and weaknesses of their work? 
3. Clarity: Is the submission clearly written? Is it well organized? Does it adequately inform the reader? 4. Significance: Are the results important? Are others (researchers or practitioners) likely to use the ideas or build on them? Does the submission address a difficult task in a better way than previous work? Does it advance the state of the art in a demonstrable way? Does it provide unique data, unique conclusions about existing data, or a unique theoretical or experimental approach?
5. Limitations: Have the authors adequately addressed the limitations and potential negative societal impact of their work? If not, please include constructive suggestions for improvement.''',

'''Given the introduction of the paper and the strengths and weaknesses you summarized, analyze the four aspects of the paper repectively: Soundness, Presentation, Contribution, and Overall Assessment. The criteria for each aspect are as follows:

A. Soundness: the soundness of the technical claims, experimental and research methodology and on whether the central claims of the paper are adequately supported with evidence. 4 excellent; 3.5; 3 good; 2.5; 2 fair; 1.5; 1 poor.

B. Presentation: the quality of the presentation. This should take into account the writing style and clarity, as well as contextualization relative to prior work. 4 excellent; 3 good; 2 fair; 1 poor.

C. Contribution: the quality of the overall contribution this paper makes to the research area being studied. Are the questions being asked important? Does the paper bring a significant originality of ideas and/or execution? Are the results valuable to share with the broader research community. 4 excellent; 3 good; 2 fair; 1 poor.

D. Overall: 
10: Award quality: Technically flawless paper with groundbreaking impact on one or more areas, with exceptionally strong evaluation, reproducibility, and resources, and no unaddressed ethical considerations.
9: Very Strong Accept: Technically flawless paper with groundbreaking impact on at least one area and excellent impact on multiple areas, with flawless evaluation, resources, and reproducibility, and no unaddressed ethical considerations.
8: Strong Accept: Technically strong paper with, with novel ideas, excellent impact on at least one area or high-to-excellent impact on multiple areas, with excellent evaluation, resources, and reproducibility, and no unaddressed ethical considerations.
7: Accept: Technically solid paper, with high impact on at least one sub-area or moderate-to-high impact on more than one area, with good-to-excellent evaluation, resources, reproducibility, and no unaddressed ethical considerations.
6: Weak Accept: Technically solid, moderate-to-high impact paper, with no major concerns with respect to evaluation, resources, reproducibility, ethical considerations.
5: Borderline accept: Technically solid paper where reasons to accept outweigh reasons to reject, e.g., limited evaluation. Please use sparingly.
4: Borderline reject: Technically solid paper where reasons to reject, e.g., limited evaluation, outweigh reasons to accept, e.g., good evaluation. Please use sparingly.
3: Reject: For instance, a paper with technical flaws, weak evaluation, inadequate reproducibility and incompletely addressed ethical considerations.
2: Strong Reject: For instance, a paper with major technical flaws, and/or poor evaluation, limited impact, poor reproducibility and mostly unaddressed ethical considerations.
1: Very Strong Reject: For instance, a paper with trivial results or unaddressed ethical considerations.''',

'''Given the introduction of the paper, the strengths and weaknesses you summarized, and your analysis of the four aspects, provide a score for each aspect accordingly. Ensure that each aspect is orthogonal to the others—for example, Presentation should not affect the evaluation of Soundness or Contribution. You are a rigorous reviewer, and the conference has an acceptance rate of no more than 30\%. 
Provide the scores in the following format:[[Soundness\_llm: X]], [[Presentation\_llm: X]], [[Contribution\_llm: X]], [[Overall\_llm: X]].'''
\end{myblock}

    \section{Dataset examples}
    \label{app:data_examples}
    \begin{myblock}{\textsc{Creative writing, Creative}}
        Sam Greenfield is a clumsy, orphaned young woman whose life has been constantly plagued by misfortune and has recently been forced out from her foster home, much to the dismay of her younger friend Hazel, who is hoping to be adopted soon. One night, after sharing a panini with a black cat, she finds a penny she hopes to give to Hazel for her collection of other lucky items to help her get adopted. The next day, Sam notices that the penny has made her luck significantly improve. However, she soon loses the penny by inadvertently flushing it down a toilet. As Sam bemoans her error, she encounters the cat again and tells him what happened, which causes the cat to berate her for losing the penny. Shocked, Sam follows the cat through a portal to the Land of Luck, where creatures like leprechauns create good luck for the people on Earth. The cat, named Bob, tells Sam he needs the penny for traveling purposes and that he will be banished if word gets out that he lost it. Bob and Sam make a deal to get another penny from the Penny depot for Hazel to use before returning it to Bob. Bob uses a button from Sam to pass off as a penny while she sneaks into the Land of Luck using clothes belonging to Bob's personal leprechaun, Gerry. Throughout the journey, Sam comes to learn how the good luck is managed by a dragon, and that bad luck is managed underneath the Land of Luck. Following a disaster at the Penny depot which causes Gerry to learn about Sam's identity, Gerry uses a drone to retrieve the missing penny on Earth but the drone gets lost in the In-Between, a space between the Good and Bad Luck lands. Sam and Bob go to the In-Between, which is managed by a unicorn named Jeff. Jeff manages a machine called the Bad Luck Apparat that keeps bad luck specks from sticking which feeds the Randomizer, another machine that sends both good and bad luck into Earth. Jeff tells the pair he found the penny and has returned it to the depot. Not deterred, Sam decides to visit the dragon in hopes to get another penny. The dragon, named Babe, shares a moment with Sam by telling her how better things would be if everyone had good luck before giving her a new penny. But Sam sacrifices her penny after Bob is caught for faking his travel penny to spare him from banishment. Still wanting to help Hazel, Sam and Bob decide to temporarily shut down the Bad Luck Apparat to prevent bad luck from going to the Randomizer and give Hazel the luck she needs to get adopted. However, the bad luck specks start to clog Jeff's machines and destroy the good luck and bad luck stones within the Randomizer, which itself brings bad luck to the Land of Luck and Earth. Seeing Hazel did not get adopted because of this, learning that Bob is actually an unlucky English cat and having been found out as a human, Sam sulks in remorse. Bob apologizes and tells Sam that Hazel is the luckiest girl for having Sam at her side. Sam realizes things can be fixed because she remembers seeing some good luck in Bad Luck land while on her way to the In-Between. Back in Bad Luck, they find it in a tiki bar where the bartender, a root monster named Rootie, who is Bob's old friend, gives them a jar of good luck they have been using. They take it to Babe to forge new good and bad luck stones. However, while Babe creates a bad luck stone, she creates two good luck stones, wanting to create a world with only good luck. Before she can place them, Sam tells Babe people need bad luck as much as good luck. Realizing her mistake, she allows Sam to place the bad luck stone, and  good luck is restored to normal, where Sam sees Hazel finally getting adopted by a new family. Bob is offered to keep his job at the Land of Luck, but decides he wants to live with Sam. One year later back on Earth, Hazel's family spends time with Sam and Bob as they have finally accepted their bad luck.
    \end{myblock}
    \begin{myblock}{\textsc{Creative writing, Uncreative}}
        Fortune's Folly is a heartwarming comedy that follows the misadventures of Lily Thompson, a young woman who seems to have been born under an unlucky star. Lily, an awkward, parentless young lady, has been evicted from her care home, leaving her junior companion, Daisy, who is eagerly awaiting adoption, heartbroken.   Lily's life has been a series of unfortunate events, from slipping on banana peels to accidentally causing minor explosions in chemistry class. Her bad luck is legendary, and she's become the town's favorite source of amusement. However, Lily's misfortune takes a turn when she stumbles upon a mysterious old woman who claims to be a fortune teller.   The fortune teller, Madame Zara, promises to change Lily's luck if she completes a series of bizarre tasks. With nothing to lose, Lily embarks on a hilarious journey, which includes wrestling a pig, singing opera in a chicken suit, and even attempting to steal the mayor's prized gnome collection.   Meanwhile, Daisy, missing her friend, decides to take matters into her own hands. She embarks on a mission to find a family willing to adopt both her and Lily. Daisy's attempts to find a family are equally hilarious, involving a series of disastrous interviews with potential parents, including a couple of circus performers, a pair of overly enthusiastic survivalists, and a family of competitive eaters.  As Lily completes each task, her luck seems to improve, leading to a series of comedic twists and turns. She accidentally discovers a hidden treasure, saves the town from a disastrous flood, and even manages to win the heart of the handsome local baker, much to the town's surprise.  In the end, Lily's journey of self-discovery and Daisy's relentless pursuit of a family lead them to a surprising revelation. The fortune teller, Madame Zara, is actually a wealthy recluse looking for a family to leave her fortune to. Touched by Lily and Daisy's bond and resilience, she decides to adopt them both, providing them with the family they've always wanted.  "Fortune's Folly" is a comedy of errors that proves that sometimes, the greatest fortunes come in the most unexpected ways. It's a story of friendship, resilience, and the hilarious lengths one will go to change their luck.
    \end{myblock}
    \begin{myblock}{\textsc{Problem-solving, Creative}}
        You need to give your stubborn dog his vital medication, but he refuses to take it orally. Sadly, you don't have pill pockets. You do have a bristly hairbrush, a loaf of multigrain bread, a plastic cup with a tight lid, a jar of creamy peanut butter, a dog leash, and a sharp kitchen knife. However, your dog, unfortunately, has a severe allergy to peanut butter. How can you tactfully administer the pill to your dog using these items?\\
        
        First: Using a kitchen knife, slice off a small portion of bread.\\
Second: Carefully push the medication into the bread piece, ensuring it's well-hidden.\\
Third: Transfer the medicated bread morsel into your plastic container.\\
Fourth: Secure the container by affixing the lid firmly.\\
Fifth: Vigorously agitate the sealed container to further conceal the pill within the bread.\\
Finally: Remove the lid and present the prepared bread treat to your canine companion for consumption.
    \end{myblock}
    \begin{myblock}{\textsc{Problem-solving, Uncreative}}
        You're outside in the rain and realize you have important papers in your backpack that need to stay dry. You can see water is already beginning to seep through your backpack. Available tools: pencil, umbrella (doesn't open properly, limited functionality), plastic bag, carabiner, credit card, big paperclip (rusted), old newspaper, shoe lace (frayed, unreliable). How can you protect your important documents from getting wet?\\

        To safeguard your documents from moisture:\\
1. Extract all papers from your backpack.\\
2. Insert the documents into a waterproof plastic container (bag).\\
3. Secure the container by creating multiple folds at the top opening to ensure a watertight seal.\\
4. Reinsert the now-protected paperwork into your backpack for transportation.
This method keeps your important documents dry by creating a waterproof barrier between them and any potential moisture that might enter your backpack.

    \end{myblock}
    \begin{myblock}{\textsc{Research ideation, Creative}}
        A general autonomous agent must be able to model the complex behaviors, goals, and beliefs of other agents in its environment to refine its decision-making. Traditionally, this task—known as opponent modeling in adversarial settings—has been approached by learning an opponent model online. However, online learning is often impractical and inefficient. In many real-world situations, such as on e-commerce platforms, it is not feasible to continuously engage with an opponent to learn their behavior; instead, one must rely on passively accumulated historical data. Moreover, even when online learning is possible, an opponent might change its policy across episodes, meaning that by the time a new model is learned, the opponent’s strategy may have already shifted. This necessitates many rollouts to acquire an updated model, making the process prohibitively time-consuming.\\
        
 To tackle these challenges, we extend the opponent modeling problem to an offline setting—referred to as Offline Opponent Modeling (OOM)—to improve sample efficiency and overall accessibility. The key idea is that obtaining replay data or historical plays of sufficient quality is generally straightforward. This offline data can be used to learn a response policy against a fixed opponent policy, bypassing the need for extensive trial-and-error exploration. Many practical applications follow this principle; for example, players master Go strategies by practicing from high-quality exercises, and basketball coaches analyze replay videos to understand the playing styles of specific opponents.\\

 Recent research has demonstrated that Transformers, when pre-trained for decision-making tasks, exhibit a robust ability to learn in-context in meta-reinforcement learning. This is exactly what is needed in opponent modeling. When deploying an opponent model in a new environment, it is essential to adapt quickly to previously unseen opponent policies without requiring gradient updates. With this in mind, we introduce Transformer Against Opponents (TAO), a Transformer-based approach designed to address the OOM problem.\\

 TAO operates in three stages. First, it employs an Opponent Policy Encoder (OPE) to transform opponent trajectories into latent representations that capture essential policy characteristics. Next, using these embeddings, an In-context Control Decoder (ICD) is pre-trained via supervised learning to learn how to respond appropriately to different opponent trajectories. Finally, in deployment, TAO adapts to unknown opponent policies in a new environment by leveraging an Opponent Context Window (OCW) that continuously collects opponent trajectories and facilitates in-context learning.\\

 Theoretically, TAO is equivalent to Bayesian posterior sampling in opponent modeling, and analyses confirm its convergence in recognizing opponent policies. We demonstrate the effectiveness of TAO through experiments in both sparse and dense reward environments, comparing it against a range of baseline methods in opponent modeling and offline meta-reinforcement learning. The results show that TAO performs exceptionally well across different settings, and ablation studies highlight the critical role of its individual components. Notably, when faced with unseen opponent policies in a new environment, TAO adapts faster and more effectively than alternative approaches, showcasing its powerful in-context learning capability.
    \end{myblock}
    \begin{myblock}{\textsc{Research ideation, Uncreative}}
        Euclidean clustering is a fundamental task with applications in robotics, computer vision, environmental modeling, and many other fields. In this setting, the goal is to identify groups of data points that are close to one another in Euclidean space under a standard distance metric. Traditional methods—such as variants of Lloyd’s algorithm or spectral clustering—approach the problem by first selecting initial cluster centroids and then repeating two main steps: assigning each data point to its nearest centroid and recalculating the centroids as the mean of the points associated with each cluster. This iterative process continues until the centroids stabilize or a fixed number of iterations is reached, with the objective of minimizing the overall sum of squared distances between data points and their centroids. A known limitation of these approaches is that their performance heavily depends on the initialization and may become trapped in local minima.\\

 In our work, we derive a closed-form solution to the clustering problem that forgoes the need for such iterative, initialization-sensitive procedures. Instead, we leverage the fact that clustered data tend to lie near a specific subspace in the feature space. The projection operator onto that subspace contains the clustering information in the form of an adjacency matrix, which can be extracted by a simple thresholding operation. Because projection operators are unique, we can determine both the subspace and the corresponding clustering through a straightforward singular value decomposition.\\

 Our main theoretical result shows that, when clusters are well defined—meaning they are sufficiently separated and each individual cluster is relatively compact—our closed-form solution is guaranteed to identify the clusters correctly with certainty. In this context, the degree of separation between clusters serves as a proxy for noise, ensuring that the method’s correctness depends only on the level of noise and not on its specific distribution. The proof of this result is both elegant and direct, relying on a few key insights and a classical theorem from matrix perturbation theory.\\

 While some practical applications may tolerate heuristic accuracy measures, other sensitive applications demand outputs with certified reliability. For instance, in single-cell sequencing—where identifying correlations between gene activity and cell types is crucial—modern methods often fall short, relying on algorithms that cannot guarantee correctness and on indirect accuracy indicators. Our approach directly addresses this shortcoming.\\

 We support our theoretical contributions with extensive experiments on both synthetic data and several real datasets from single-cell sequencing. These experiments not only validate our analysis but also demonstrate that simple adaptations of our closed-form solution yield practical algorithms that gracefully handle increasing noise. In many cases, these variants rival or even exceed the current state-of-the-art in both accuracy and speed.\\

 The remainder of the paper is organized as follows. Section 2 outlines the problem and provides an overview of related work. Section 3 presents our main contributions, while Section 4 details the key ideas and proofs behind our theoretical findings. Section 5 describes practical adaptations of our solution, and Section 6 presents our experimental results.

    \end{myblock}

\end{document}